# CLASSIFICATION OF QUASARS, GALAXIES, AND STARS USING MULTI-MODAL DEEP LEARNING

Sabeesh Ethiraj
CH Bangalore, India
sabeesh90@yahoo.co.uk

Bharath Kumar Bolla
Verizon, India
bolla111@gmail.com

*Abstract*— The universe is the vast expanse of cosmic space consisting of billions of galaxies. Galaxies are made of billions of stars that revolve around a gravitation center of the black hole. Quasars are quasi-stellar object which emits electromagnetic radiation more potent than the luminosities of the galaxies, combined. In this paper, the fourth version, the Sloan Digital Sky Survey (SDSS-4), Data Release 16 dataset, was used to classify the SDSS dataset into galaxies, stars, and quasars using machine learning and deep learning architectures. We efficiently utilize both image and metadata in tabular format to build a novel multi-modal architecture and achieve state-of-the-art results. For the tabular data, we compared classical machine learning algorithms (Logistic Regression, Random Forest, Decision Trees, Adaboost, LightGBM, etc.) with artificial neural networks. Deep learning architecture such as Resnet50, VGG16, EfficientNetB2, Xception, and Densenet121 have been used for images. Our works shed new light on multi-modal deep learning with their ability to handle imbalanced class datasets. The multi-modal architecture further resulted in higher metrics (accuracy, precision, recall, F1 score) than models using only images or tabular data.

## I. INTRODUCTION

The universe is the vast expanse of cosmic space consisting of billions of galaxies. Galaxies are made of billions of stars that revolve around a gravitation center of the black hole. Quasars are quasi-stellar object which emits electromagnetic radiation more potent than the luminosities of the galaxies, combined. There have been numerous large-scale surveys that have been done to map the universe and the celestial objects present in it. The most important surveys are the Sloan Digital Sky Survey (SDSS) (Blanton et al., 2017), which commenced observations of the universe in 1998. There have been four major phases of this survey with multiple data releases. The information captured by the SDSS survey includes optical, spectroscopic, and photometric information, along with an array of other observations. The current ongoing and the latest phase of this survey is the SDSS-4 survey (Data Release 16), which has been functional and consists of observations recorded from August 2018 till the present. The paper aims to study and evaluate multi-modal deep learning architecture and evaluate its performance compared to baseline machine learning and deep learning models in the Sloan Digital Sky Survey classification. This will help astronomers and scientists identify and classify the 3 million celestial objects in an automated and efficient manner. The same can also be extended in the classification of other large-scale astronomical datasets in the mapping of the universe. The objective of this paper is as follows:

1) To classify the SDSS dataset into galaxies, stars, and quasars by applying machine learning models on tabular data and compare the performance with Artificial Neural Networks

2) To classify the SDSS dataset into galaxies, stars, and quasars by applying deep learning models on image data.

3) To apply multi-modal deep learning architecture on image and tabular data and compare the performance with unimodal architectures

## II. RELATED WORK

Many Large Surveys of the universe have been done over for a while. Amongst the most popular surveys which capture information about the celestial objects in the universe is the Sloan Digital Sky Survey (SDSS), which consists of photometric and spectroscopic information about 3 million astronomical objects such as galaxies, stars, and quasars in both tabular and image format. Machine learning and Deep learning architectures are being continually designed and utilized in many large-scale astronomical surveys. Skynet [1] is one such neural network used in regression, classification, and clustering algorithms. A similar neural network called AstroNN [2] was designed and built specifically for astronomical surveys used to analyze spectroscopic data of the APO - Apache Point Observatory - Galactic Evolution Experiment. Algorithms such as SKYNET that are based on Artificial neural networks have also been used in Astronomical datasets. The work done by [3] demonstrates the use of this network in astronomical classification and its application in creating the BAMBI algorithm for accelerated Bayesian inference. The application of Skynet has also been used in the classification problem of gamma ray bursts.

### A. Machine Learning on SDSS Dataset

The other spectrum of research involves the classification of these astronomical objects into their corresponding classes. The work done by Acharya et al. [4] in 2018 wherein the entire SDSS-3, DR-12 dataset was classified. The classification was built on the same photometric parameters you, g, i, r, and z using PySpark on Google Proc. This was a significant attempt to classify the entire dataset wherein both binary, and multi-class classification was done. This study was later followed by a comparative evaluation study in 2020 by Petrusevich [5] on the SDSS-4 DR 14 dataset. Baseline machine learning models such as Logistic regression, Naive Bayes Classifier, Gradient Boosting, Decision Trees, and Random Forest were applied to this dataset and the enhanced version using feature engineering techniques. It was shown that these baseline machine learning models performed better than or as good as conventional deep learning models in terms of Accuracy, Precision, and Recall on both the baseline and enhanced dataset. Similar machine learning models were also used in the classification of images, as is the work done by du Buisson et al. in 2015 [6], where different images were created from the actual images of the sky at two different points in time on the transient images of the SDSS 2- survey. Deep learning models such as Skynet [1], which were designed explicitly for astronomical datasets, were used in the classification algorithm.

### B. Deep Learning on Astronomical Classification

Images created using photometric parameters of the observatories were later used in building customized novel (Convolutional Neural Networks) CNN architecture using temporal convolution and filter convolutions that would classify data with very high accuracy[7]. It was also shown that combining these models with a conventional Machine Learning model such as K-Nearest Neighbors and Random Forest Classifier would result in a complementary effect resulting in an increased performance of these models. Similar work by Khramtsov et al. in 2019 [8] was done on the SDSS DR 9 dataset, and Galaxy Zoo2 dataset, wherein images were built using similar techniques as mentioned above from the photometric parameters. The Deep Network Xception [9] was used to build the classification algorithm that performed better in combination with an SVM.

### C. Multi-Modal Deep Learning

With the evolution of CNNs and increasing evidence of better model performance due to the complementary effect of various machine learning models, focus shifted on combining various modalities and models to incorporate multi-modal learning in these classification problems. These multi-modal techniques are predominantly being used in the health care domain. The work done by Vaghefi et al. in 2020 [10]in the classification and diagnosis of Age-Related Macular Degeneration uses a combination of CNN modalities such as Optical Coherence Tomography, Optical Coherence Tomography – Angiography, and Color Fundus Photography. Models were built using the Resnet-Inception-V2 design. The multi-modal CNN performed better than a single CNN architecture or a bi-modal architecture. Similar work was done by Le et al. [11]in 2017 in the diagnosis and the classification of Cancerous / Non-Cancerous lesions of prostate and clinically significant/indolent Prostate Cancer. Two modalities using Apparent Diffuse Coefficient (ADC) and T2 weighted images combined with CNN architecture trained on Resnet 50, GoogleNet 22, and VGG 16 and the performance of Resnet 50 in the classification algorithm was found to be superior to all the other models in terms of accuracy, specificity, and sensitivity. Pre-trained weights in networks such as ResNet/ GoogleNet do not need to consistently perform better than customized deep learning models with lesser layers. This has been evident in work done by Song et al., 2019 [12], who proposed a multi-modal architecture in predicting the temperature on ten different zones in the casting of the steel. Latin hypercube method was used to eliminate the bias in the dataset.

### D. Fusion Techniques in Multi-Modal Architectures

Different kinds of fusion techniques have been used while combining modalities in deep learning. The work done by [10] Vaghefi et al., 2020 used a Resnet-Inception-V2 architecture where infusion was done after training the three different modalities separately and then fusing them at the level of the Max Pooling layer and then feeding them into the Resnet Architecture. Other kinds of fusion such as merging fully connected layers, element-wise summation of feature maps, deduction in the dimensionality of the merging vectors using PCA to compensate for the over-representation of a particular modality [11] were also done. The performance of the multi-modal architecture also depends upon the level at which the fusion occurs. Conventional models use late fusion techniques where the fusion of the models occurs before the softmax layer. However, the work done by Xu et al. [13]on the identification of cervical dysplasia using two modalities, namely cervigram and the clinical reports of PAP test and HPV showed that early fusion in the multi-modal architecture (built using Alexnet) resulted in better model metrics. The hypothesis was that early fusion would extract features better than late fusion techniques if a tighter correlation existed between the features. A similar work done by Liu et al. in 2018 [14] evaluated the multiplicative network fusion method in multi-modal architecture over the conventional additive method. It was found that the multiplicative method coupled with boosting extension, which penalized only the wrong classes for misclassification, performed better than the additive method. Four different versions of the Resnet architecture, namely Resnet-base, Resnet-Add, Resnet-Mul, and Resnet-Mulmix (built on Resnet 18, Resnet 32, and Resnet 110), were used here. The advantage of early fusion has also been validated in the diagnosis of MRI brain depicting Alzheimer's disease [15], wherein intermediate layers of the CNN architecture were extracted and the clinical data of the patients were added using the Gram Matrix method.

## III. METHODOLOGY

Sloan Digital Sky Survey (SDSS) is an extensive survey of the universe to map and identify over 3 million astronomical objects present. The survey captures large-scale patterns of galaxies, stars, and other celestial bodies. The SDSS was commissioned in 1998 and surveying the universe using the telescopes located at Apache Point Observatory. The dataset used in this application is SDSS – IV DR 16. SDSS-4, Data release 16 is the 16th data release of the fourth phase of the SDSS sky survey which has been capturing observations through August 2018 till date. The survey also consists of cumulative data from the previous releases as well. The DR 16 release consists of six types of data namely, images, optical-spectra, infrared-spectra (APOGEE/APOGEE-2), IFU spectra (MaNGA), stellar library spectra (MaStar) and catalog data (parameters such as magnitudes and redshifts obtained from spectra). The dataset used in this research has tabular data (catalogue data, derived from images, spectra as mentioned above) and image data.

The tabular data has been downloaded using a customized SQL query from the SDSS website through which the SDSS server can be accessed. Image data consists of the images obtained using the APIs provided in the SDSS site. The image data is fetched using two parameters (r[pods nm 0a and dec) that correspond to the image's location in the sky. The images are obtained with a scale ranging from 0.01 onwards to 0.1 using the coordinates ra and dec. Ra and Dec represent the location of the object in the sky. A scale of 0.1 would focus on the center of the image, which is a zoomed-in picture of the target body in the image. This would also exclude other artifacts, thereby increasing accuracy at the time of analysis and model building. The dimension of the images downloaded is in the pixel range of 2048 x 2048. This is to capture the maximum possible information from the features at convolution for model building. A subset of the total dataset, which consists of 1000 data points, including images and the tabular data, is used in this study. The data is split into a ratio of 70:30 into the train and the validation set.

*A. Data Preprocessing of Tabular Data*

The tabular data consists of six dependent variables and one target variable. The following pre-processing steps have been done to check for null values, outlier, and class imbalance. All the six variables present here are continuous numerical variables, and no encoding is required to convert them. Exploratory data analysis (EDA)has been done on the dataset to identify outliers. The target variable is a categorical variable with three different types of categories. Label encoding of the target variable has been done. Class Imbalance of the target variables has been checked, and appropriate weighting has been applied to that class during model training. Standardization of the six feature variables has been done before passing them into the machine learning algorithms, including Artificial Neural networks. Standardization is subtracting each value of the feature set from its corresponding mean divided by the standard deviation. This would distribute the data in a standardized fashion around the mean.

*B. Data Preprocessing of Image Data*

The image data downloaded using the API from the SDSS website can range from 64 pixels to 2048 pixels. However, the target object in this data corresponds to the ra and dec coordinates centered on the image. Hence, zooming the image is done when downloading the dataset to isolate any artifacts present in the image, which may affect the learning algorithms. The image has a default zoom factor of 0.1. The input images are then resized to 512 x 512 pixels. The images are also normalised before being fed into the deep networks. Additionally, non-normalized images are also tested to study their effect on the accuracy of the model.

*C. Machine Learning Models*

Machine learning models have been evaluated using Pycaret Library. The evaluation of the models has been done using overall accuracy, recall, precision, and F1 Score. Artificial Neural Network has been built using three dense layers.

*D. Multi-Modal Deep Learning Architecture*

A Multi-modal architecture has been built using which combines two modalities - Images and tabular data. The tabular data is passed through an Artificial Neural Network. The ANN network has been tuned during unimodal training, and the best model has been used to merge the output features with that of the features from the transfer learning CNN architectures. Transfer learning architectures such as Resnet50 Xception and EfficientNetB2 have been used to build the multi-modal architectures. Functional APIs of Tensorflow-Keras has been used to create functional models wherein the output of the last convolutional layer is extracted, converted to a single-dimensional vector and concatenated with the single-dimensional vector output of the ANN. Tuning of the number of feature representations in the last layer of ANN and the CNN has been done to identify the best possible concatenation layer at the level of feature fusion. The concept of late fusion technique has been implemented in this study to fuse features from both modalities. The output is later passed through series of Fully Connected layers or directly to a softmax layer to give the desired output via a softmax function. The performance of this model is expected to dominate the unimodal architectures. A representation of the multi-modal architecture is as shown in Fig.3

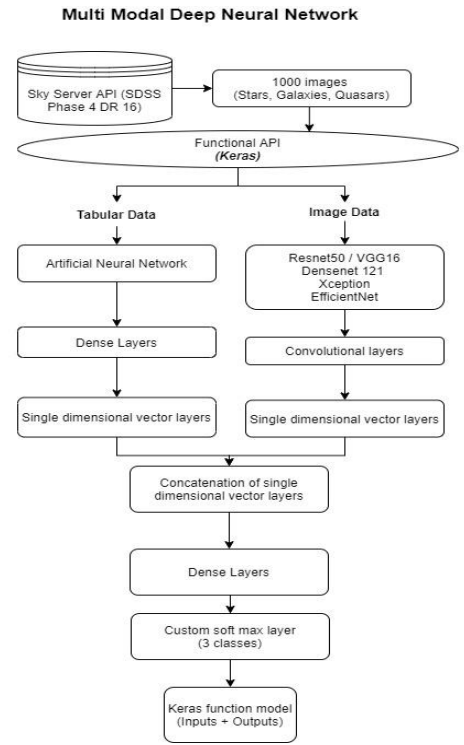

*Fig 1. Multimodal architecture*

*E. Loss Functions*

The loss functions used in various models such as Logistic Regression, ANNs, and CNN models are Categorical Cross-Entropy loss as this is a multi-class classification problem. The following equation defines categorical cross entropy loss.

$$CE = -\sum_{i}^{C} t_i log(s_i)$$

*Equation 1. Cross Entropy Loss*

*F. Evaluation Metrics*

The evaluation metrics used in this study are Recall, Accuracy, Precision and F1 score.

## IV. ANALYSIS

*A. Tabular data*

On initial analysis of tabular data, it is found that there exists a class imbalance in the dataset w.r.t the Quasar class.

The distribution of the data points in the dataset is shown below in Fig 4 and Table 1. Appropriate class imbalancing techniques such as the incorporation of class weights have been used to nullify the effect of class imbalance.

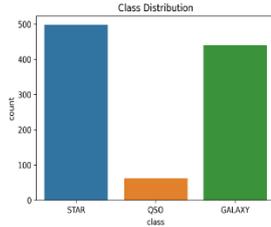

*Fig 2. Class imbalance*

*Table 1. Distribution of Classes*

| Class | Total | % of Distribution |
|---|---|---|
| Galaxy | 440 | 44.0 |
| QSO | 62 | 6.2 |
| Star | 498 | 49.8 |

### B. Image Data

The shape of the original images is 2048x2048 pixels which have been resized into 512x512 pixels before passing into the input layer. It is observed that quasar and a star have a similar appearance in terms of pixels and may affect model training due to the less distribution of the quasar class resulting from class imbalance. The visualizations of the three different classes, galaxy, star, and quasar, are shown below in Fig 5.

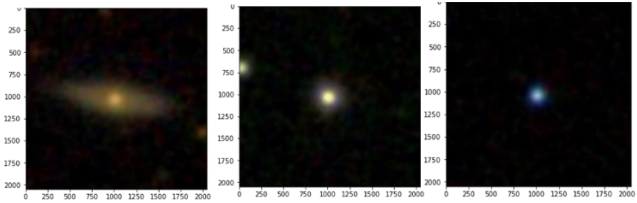

*Fig 3. Galaxy, Quasars, and Stars*

### C. Deep Learning Architectures

Deep learning models are built using transfer learning by *gradually increasing the number of training layers* in the pre-trained architecture until the ideal model is identified in validation accuracy. The summary of the architectures is shown below in Table 2

*Table 2. Transfer learning Parameters and Layers*

| Transfer Learning Model | Total layers | No of parameters |
|---|---|---|
| Resnet50 | 176 | 23,593,859 |
| Xception | 133 | 20,867,627 |
| VGG16 | 20 | 14,716,227 |
| EfficientNet B2 | 340 | 7,772,796 |
| DenseNet121 | 428 | 7,040,579 |

Resnet50 has the maximum number of parameters, while DenseNet121 has the least number of parameters.

### D. Multimodal Architecture

Multi-modal architectures are built by fusing the best performing transfer learning moels and the ANN architecture trained on tabular data. The number of trainable layers differs from the initially tuned transfer learning models because the tuning is done along with ANN networks. Hence, the ideal number of trainable layers may vary. The summary of the architecture is shown below in table 3.

*Table 3. Trainable / Non Trainable layers in Multi-modal architecture*

| Model | Total layer | Non trainable layers | Trainable Layers | No of parameter |
|---|---|---|---|---|
| Resnet50 | 176 | 150 | 16 | 24,321,519 |
| Xception | 133 | 125 | 8 | 22,131,607 |
| EfficientNet B2 | 340 | 330 | 10 | 8,764,776 |

## V. RESULTS

### A. Machine Learning on Tabular Data

*Table 4. Machine Learning Models Experiments*

| Model | Acc | AUC | Rec | Pre. | F1 |
|---|---|---|---|---|---|
| **Extra Trees Classifier** | **0.96** | **0.98** | **0.92** | **0.96** | **0.96** |
| LGBM | 0.96 | 0.98 | 0.92 | 0.96 | 0.96 |
| Logistic Regression | 0.95 | 0.98 | 0.93 | 0.96 | 0.95 |
| Random Forest | 0.95 | 0.98 | 0.93 | 0.96 | 0.95 |
| QDA | 0.95 | 0.98 | 0.91 | 0.96 | 0.95 |
| Gradient Boosting Classifier | 0.95 | 0.98 | 0.921 | 0.96 | 0.95 |
| SVM | 0.95 | NA | 0.93 | 0.968 | 0.95 |
| K Neighbours Classifier | 0.94 | 0.98 | 0.91 | 0.95 | 0.95 |
| Ridge Classifier | 0.94 | NA | 0.93 | 0.96 | 0.95 |
| LDA | 0.93 | 0.98 | 0.93 | 0.96 | 0.94 |
| Decision Tree | 0.92 | 0.94 | 0.87 | 0.93 | 0.92 |
| Ada Boost | 0.83 | 0.92 | 0.81 | 0.87 | 0.84 |
| Naive Bayes | 0.66 | 0.84 | 0.73 | 0.74 | 0.67 |
| **ANN** | **0.97** | **NA** | **0.97** | **0.97** | **0.97** |

From Table 4, it is evident that all *tree-based classifiers produce high accuracies*. Random Forest, Extra Tree Classifier, QDA, and Logistic Regression give high evaluation metrics in terms of accuracy, recall, and F1 scores. However, ANNs perform better than most baseline machine learning models having the *highest F1 score* among the baseline machine learning models. F1 score indicates a model's overall performance, and this high score indicates that deep networks perform better than baseline models in the case of Class imbalanced datasets.

### B. Deep Learning on Image Data

Among the deep learning models, it has been observed that newer architectures such as EfficentNetB2 and Xception architectures perform better than older architectures such as Resnet50, VGG16, and DenseNet121 on minimal tuning trainable layers in terms of accuracies and F1 scores.

Inference on Accuracies and Training Curves

It is seen that EfficeintNetB2 and Xception have higher overall accuracies (91%) and weighted Recall and Weighted Precision scores than the older architectures. Also, as seen in the training curves in Fig 4 to Fig 8, lesser overfitting is seen in the Xception and EfficientNet as inferred from the decreased train test gap. Further, higher accuracies are reached at a minimal number of epochs, as seen from the steeper curves in the newer architectures. The evaluation metrics are shown in Table 5

*Table 5. Model Evaluation Metrics*

| Model | Accuracies | Rec | Prec |
|---|---|---|---|
| Resnet50 | 90% | 90% | 89% |
| VGG16 | 89% | 89% | 89% |
| DenseNet121 | 81% | 81% | 85% |
| **EfficientNetB2** | **91%** | **91%** | **91%** |
| **Xception** | **91%** | **91%** | **91%** |

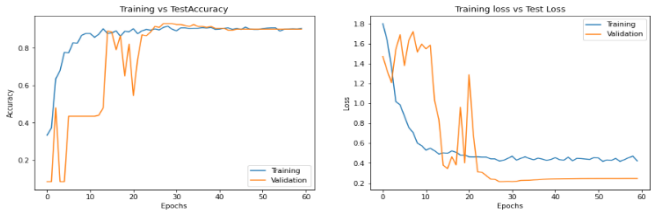

*Fig 4. Resnet Training curves*

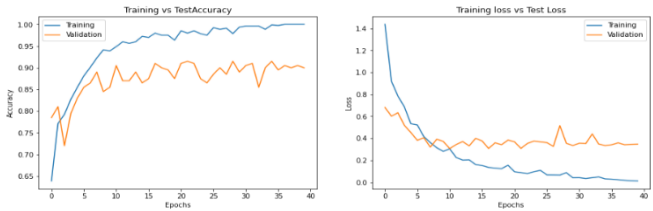

*Fig 5. VGG16 Training curves*

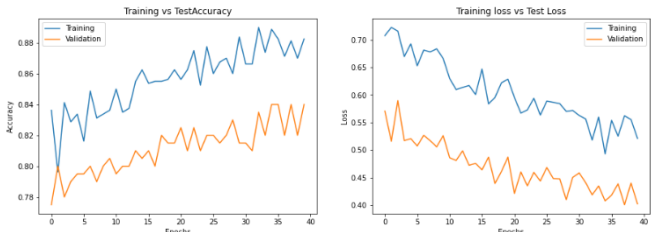

*Fig 6. DenseNet121 training curves*

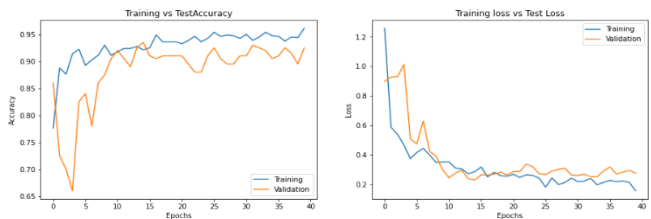

*Fig 7. Xception Training Curves*

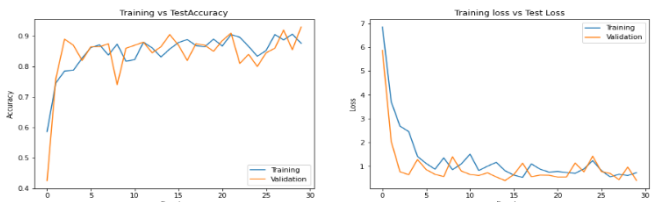

*Fig 8. EfficientNetB2 Training Curves*

Inference on Class based F1 scores

*Table 6. Evaluation of Class based F1 scores*

| Model | Q-F1 | Galaxy-F1 | Stars-F1 | Overall F1 |
|---|---|---|---|---|
| Resnet50 | 27% | 92% | 89% | 89% |
| VGG16 | **62%** | 94% | 89% | 89% |
| DenseNet121 | 46% | 92% | 78% | 82% |
| EfficientNetB2 | **63%** | 96% | 90% | **91%** |
| Xception | 59% | 98% | 89% | **91%** |

The performance of the models can be further evaluated from the class-based F1 scores, as shown in Table 6. Since this is a class imbalanced dataset, the performance of the models on these individual classes will give us a better understanding of the ideal architecture. From table 6, it is evident that **EfficientNetB2 (91%) and Xception (91%)** have the highest overall F1 scores. However, the F1 scores on the **Quasar class** (Least distributed class) are the **highest on EfficientNetB2 and VGG16** architectures.

### C. Supremacy of Multi-Modal Architectures

Three models from the above transfer learning architectures, such as Resnet50, Xception, and EfficientNetB2, have fused with ANN network trained on tabular data. Multi-modal architectures outperform unimodal architectures on all evaluation metrics without the use of any augmentation techniques. The summary of the performance of the models is shown below in Table 7

*Table 7. Multimodal Architecture Performance - Evaluation Metrics*

| Multimodal Architecture | Acc (%) | Precision (%) | Rec (%) | F1 (%) |
|---|---|---|---|---|
| Resnet50 + ANN | 98.50 | 98.63 | 98.5 | 98.53 |
| Xception + ANN | **99** | 99.02 | 99 | 98.96 |
| EfficientNetB2 + ANN | 98.5 | 98.53 | 98.5 | 98.4 |

As seen in Table 7, there is a stark difference in the accuracies of the multi-modal architectures over the unimodal transfer learning/machine learning models. In terms of **overall accuracies and F1 score, Xception Network+ANN** performs the best with the highest accuracy of 99% and F1 score of 98.96%. However, considering the number of parameters required to reach this high accuracy, **EfficientNetB2+ANN** may be ranked superior to all the models.

*Table 8. Class based F1 Score evaluation*

| Model | Q-F1 | Galaxy-F1 | Stars-F1 |
|---|---|---|---|
| Resnet50 + ANN | **92.86%** | **99.46%** | **98.4%** |
| Xception + ANN | 90% | 100% | 98.95% |
| EfficientNetB2 + ANN | 85.7% | 99.49% | 98.9% |

Inference on Class-Based F1 scores

In terms of the performance of the models on the individual classes, multi-modal architecture dominates the evaluation metrics, especially in the Quasar class as seen in Table 8. The F1 score for the Quasar class ranges between 85.7% and 92.86%, which is higher than any unimodal deep learning architecture. The performance of the models on the **Quasar class** is summarized below.

Resnet50+ANN (92.86%) > Xception+ANN (90%) > EfficientNetB2+ANN(85.7%) > Unimodal EfficientNetB2 (63%) > Unimodal VGG16(625)

Newer architectures outperform Resnet50+ANN in the other classes, according to the results. Training Curves of Multi-modal architecture

The training curves of the multi-modal architecture are shown below in Fig 9 to Fig 11. The curves show a better training patter than the unimodal deep learning with **steeper training curves and lesser oscillation.** The curves also reach the highest accuracies in less than 15 epochs, while unimodal architectures reach similar accuracies in a higher number of epochs.

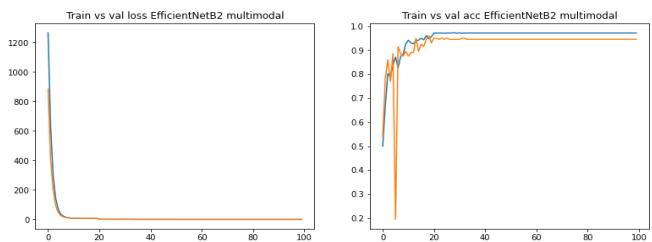

*Fig 9. EfficientNetB2 + ANN*

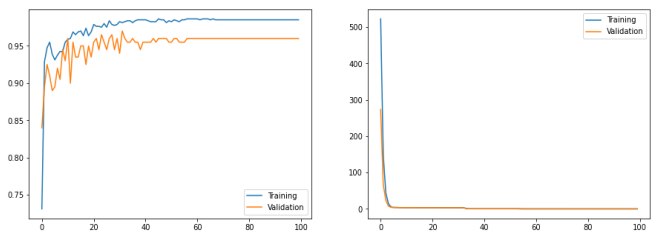

*Fig 10. Resnet + ANN*

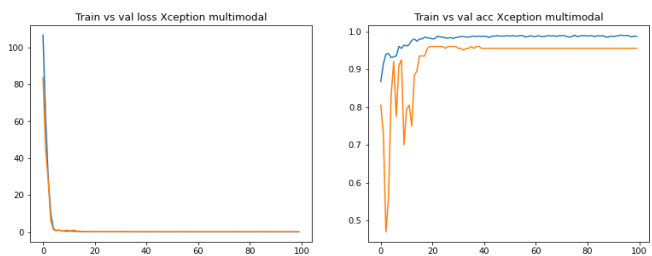

*Fig 11. Xception + ANN*

## VI. CONCLUSION

Machine learning and deep learning algorithms have become inherent components in analyzing large-scale datasets due to the complexity of the data captured. Among machine learning models that were built on tabular data, **ANNs performed better than conventional baseline models**. Further, the combination of modalities in the training algorithms increases the performance of the models in terms of accuracy, precision, and F1 scores. These **multi-modal architectures** also help alleviate the **class imbalance** problem in such datasets, as seen by their performance on the less distributed class. While multi-modal deep learning built on newer architectures outperformed older architectures in terms of overall model accuracy, both older and newer architectures performed equally well on less distributed classes, yielding high F1 scores than unimodal deep learning. However, the newer architectures may be preferred to older ones due to fewer training parameters.

To summarize, the paper establishes the need for multi-modal deep learning models and their superiority over conventional models. Specifically, the literature shows that machine learning models and deep learning models have been used independently to classify the SDSS-4 dataset. Our work is perhaps one of the initial that multi-modal architectures have experimented upon this dataset which has shown promising results in this classification scenario. Furthermore, with the release of the SDSS-V dataset soon and other large-scale observatories being commissioned to map and capture various astronomical objects, multi-modal deep learning can be used to automate such complex tasks.